\documentclass[letterpaper]{article} 
\usepackage[preprint]{aaai2027}  
\usepackage[hyphens]{url}  
\usepackage{graphicx} 
\usepackage{natbib}  
\usepackage{caption} 
\usepackage{algorithm}
\usepackage{algorithmic}

\usepackage{amsmath}
\usepackage{amssymb}

\usepackage{newfloat}
\usepackage{listings}
\DeclareCaptionStyle{ruled}{labelfont=normalfont,labelsep=colon,strut=off} 
\floatstyle{ruled}
\newfloat{listing}{tb}{lst}{}
\floatname{listing}{Listing}

\usepackage{booktabs}

\title{RoMAN-Flow: Taming Autoregressive Normalizing Flows for \\Offline Reinforcement Learning in Robotic Manipulation}
\author {
    Shaoxuan Wang\textsuperscript{\rm 1}\equalcontrib,
    Guangting Zheng\textsuperscript{\rm 1}\equalcontrib,
    Rui Huang\textsuperscript{\rm 1},
    Zhipeng Tang\textsuperscript{\rm 1},
    Sha Zhang\textsuperscript{\rm 2},
    \\Jiajun Deng\textsuperscript{\rm 1},
    Yanyong Zhang\textsuperscript{\rm 1}\corresponding
}
\affiliations {
    \textsuperscript{\rm 1}University of Science and Technology of China,\  
    \textsuperscript{\rm 2}The Chinese University of Hong Kong\\
    \{wangshaoxuan, zgt, huangrui2002, tangzhipeng\}@mail.ustc.edu.cn,\ zhangsha2048@gmail.com, \\ \{dengjj, yanyongz\}@ustc.edu.cn
}

\begin{document}

\maketitle

\begin{abstract}
Offline reinforcement learning improves robotic policies using previously collected data without further environment interaction. Yet prevalent diffusion- and flow-matching robot policies lack tractable likelihoods, limiting their use in likelihood-based offline RL post-training. AR-NFs offer both expressive action modeling and exact likelihood evaluation, but their sequential sampling incurs substantial sampling overhead during policy optimization and deployment. We present RoMAN-Flow (Robotic Manipulation with Autoregressive Normalizing Flows), an offline reinforcement learning framework that makes AR-NF policies practical for robotic manipulation by addressing this sampling bottleneck in both stages. During policy optimization, RoMAN-Flow employs a sampling-free, advantage-weighted likelihood objective that assigns higher likelihood to high-advantage actions from the offline dataset without sampling from the autoregressive policy. For efficient deployment, it distills the optimized autoregressive policy into a one-step action generator, enabling low-latency action prediction. Experiments across multiple simulated manipulation benchmarks and real-world robotic platforms demonstrate that RoMAN-Flow achieves competitive policy performance while substantially reducing inference latency. Code is available at \url{https://github.com/konnyaku28/RoMAN-Flow}.
\end{abstract}

\section{Introduction}

Vision-language-action (VLA) models have shown strong potential for robotic manipulation, yet pretrained policies often require further adaptation to downstream tasks.
Offline reinforcement learning (RL) provides a practical approach to robotic policy post-training by improving policies from previously collected reward-labeled data without additional environment interaction.

Likelihood-based offline policy optimization has recently shown strong effectiveness in large language model~\cite{gpt4,llama} post-training, where autoregressive models provide tractable exact log-likelihoods for direct optimization on fixed datasets~\cite{dpo,dro}.
In robotic manipulation, however, this paradigm remains comparatively underexplored because the dominant diffusion~\cite{chi2025diffusion} and flow-matching~\cite{lipman2022flow} policies do not readily expose low-cost exact conditional action likelihoods.
Autoregressive policies~\cite{kim2024openvla,zitkovich2023rt} provide an alternative with tractable likelihoods, but they typically convert continuous robot actions into discrete token sequences.
Such discretization introduces quantization error and may compromise fine-grained control by obscuring the continuous geometry and temporal coordination of robotic actions.
Recent transformer-based autoregressive normalizing flows (AR-NFs)~\cite{t_naf,iaf,maf} have demonstrated strong generative modeling capability in image generation~\cite{tarflow,starflow,farmer,simflow}.
By combining autoregressive dependency modeling with invertible transformations, AR-NFs substantially improve the expressiveness of conventional normalizing flows~\cite{kobyzev2020normalizing,nvp,nice,glow,kolesnikov2024jet} while retaining tractable exact likelihood evaluation.
These advances motivate us to investigate AR-NFs as a policy representation for likelihood-based offline robotic reinforcement learning.

Despite these attractive properties, the autoregressive inverse of AR-NFs makes action generation sequential, introducing substantial sampling overhead during both policy optimization and deployment.
In this work, we present \textbf{RoMAN-Flow} (\textbf{Ro}botic \textbf{M}anipulation with \textbf{A}utoregressive \textbf{N}ormalizing Flows), an offline robotic reinforcement learning framework that addresses these two efficiency bottlenecks from the policy-optimization and inference perspectives.

During offline RL post-training, RoMAN-Flow builds upon Implicit Q-Learning (IQL)~\cite{kostrikov2021offline}, whose actor update performs advantage-weighted maximum likelihood over actions contained in the offline dataset.
Because this update does not require actions sampled from the current policy, the AR-NF only applies its parallel forward transformation to evaluate the exact likelihoods of dataset action chunks, avoiding costly autoregressive reverse sampling during reinforcement learning.
RoMAN-Flow therefore directly increases the likelihood of high-value offline actions while retaining the expressive continuous action distribution modeled by the AR-NF policy.

To accelerate inference, we adapt one-step distillation~\cite{farmer,biflow} to compress the trained RoMAN-Flow policy into a one-step actor. By matching the post-trained teacher's intermediate flow states and final action chunks, the one-step student learns to approximate its inverse transformation while generating all actions in parallel. The distilled policy largely preserves the teacher's task performance and substantially reduces inference latency by eliminating the sequential autoregressive reverse process.

We evaluate RoMAN-Flow on LIBERO~\cite{liu2023libero}, MetaWorld~\cite{yu2020meta}, RoboMimic~\cite{mandlekar2021matters}, and real-world robotic manipulation tasks.
Across these settings, RoMAN-Flow achieves competitive policy performance while reducing policy inference latency by nearly an order of magnitude.
These results demonstrate that modern autoregressive normalizing flows provide a practical alternative to diffusion and flow-matching policies~\cite{chi2025diffusion, black2024pi0, physicalintelligence2025pi05} for offline robotic policy post-training.

Our main contributions are summarized as follows:

\begin{itemize}
\item An offline reinforcement learning framework for robot policy learning, namely RoMAN-Flow, that enables expressive continuous-action modeling while retaining tractable, exact likelihood evaluation through autoregressive normalizing flows (AR-NFs).

\item 
A sampling-free NF-IQL procedure that performs advantage-weighted likelihood optimization exclusively on offline dataset actions, avoiding costly autoregressive policy sampling during post traing.

\item 
A one-step policy distillation to eliminate autoregressive inference overhead while preserving policy performance, and validate the framework across multiple simulated benchmarks and real-world robotic tasks.
\end{itemize}
\section{Related Work}

\subsection{Generative Policies for Robot Learning}

Generative policies have become a dominant approach to robotic manipulation because they can represent complex and multimodal action distributions. Diffusion Policy introduced conditional diffusion models for visuomotor control~\cite{chi2025diffusion}, followed by extensions to 3D observations and vision-language-action modeling~\cite{ze20243d,wen2025tinyvla}. Flow matching provides an alternative generative formulation based on conditional vector fields~\cite{lipman2022flow} and has been adopted by large-scale VLA models such as $\pi_0$~\cite{black2024pi0} and $\pi_{0.5}$~\cite{physicalintelligence2025pi05}. However, diffusion and flow-matching policies do not expose low-cost exact action likelihoods: diffusion models rely on denoising objectives, while continuous flows require density tracking along the generation ODE~\cite{ho2020denoising,song2020score,chen2018neural,grathwohl2018ffjord}. Their offline RL post-training therefore typically requires surrogate or specialized optimization procedures.

\subsection{Offline RL and Normalizing-Flow Policies}

Offline RL improves policies using fixed reward-labeled datasets while controlling distribution shift. Representative methods include behavior-regularized optimization~\cite{fujimoto2021minimalist} and advantage-weighted policy extraction~\cite{peng2019advantage,nair2020awac,kostrikov2021offline}. Likelihood-based offline policy optimization has also shown strong effectiveness in large language model post-training, where autoregressive models permit direct evaluation of policy likelihoods~\cite{dpo,dro}. For diffusion policies, Diffusion-QL introduces value guidance during denoising~\cite{wang2022diffusion}, EDP adapts diffusion training to established offline RL objectives~\cite{kang2023efficient}, and IDQL performs critic-guided extraction from a diffusion behavior model~\cite{hansen2023idql}. These methods avoid direct exact-likelihood optimization through denoising surrogates, value guidance, or sampling-based extraction.

Normalizing flows~\cite{kobyzev2020normalizing,papamakarios2021normalizing,mathieu2020riemannian} instead provide exact density evaluation through invertible transformations~\cite{vnf,nvp}, making them naturally compatible with likelihood-based policy optimization. Although conventional normalizing flows have historically been limited in expressiveness, recent autoregressive normalizing flows, including TarFlow~\cite{tarflow}, STARFlow~\cite{starflow}, and SimFlow~\cite{simflow}, demonstrate substantially improved generative modeling while retaining exact likelihoods. Their autoregressive inversion, however, leads to slow generation; FARMER~\cite{farmer} and BiFlow~\cite{biflow} address a similar limitation through a learned one-step inverse. Building on these advances, we study AR-NFs as robotic policies, post-train them through likelihood-based offline RL, and distill the optimized policy for efficient deployment.

\section{Method}

\begin{figure*}[t]
\centering
\includegraphics[width=\textwidth]{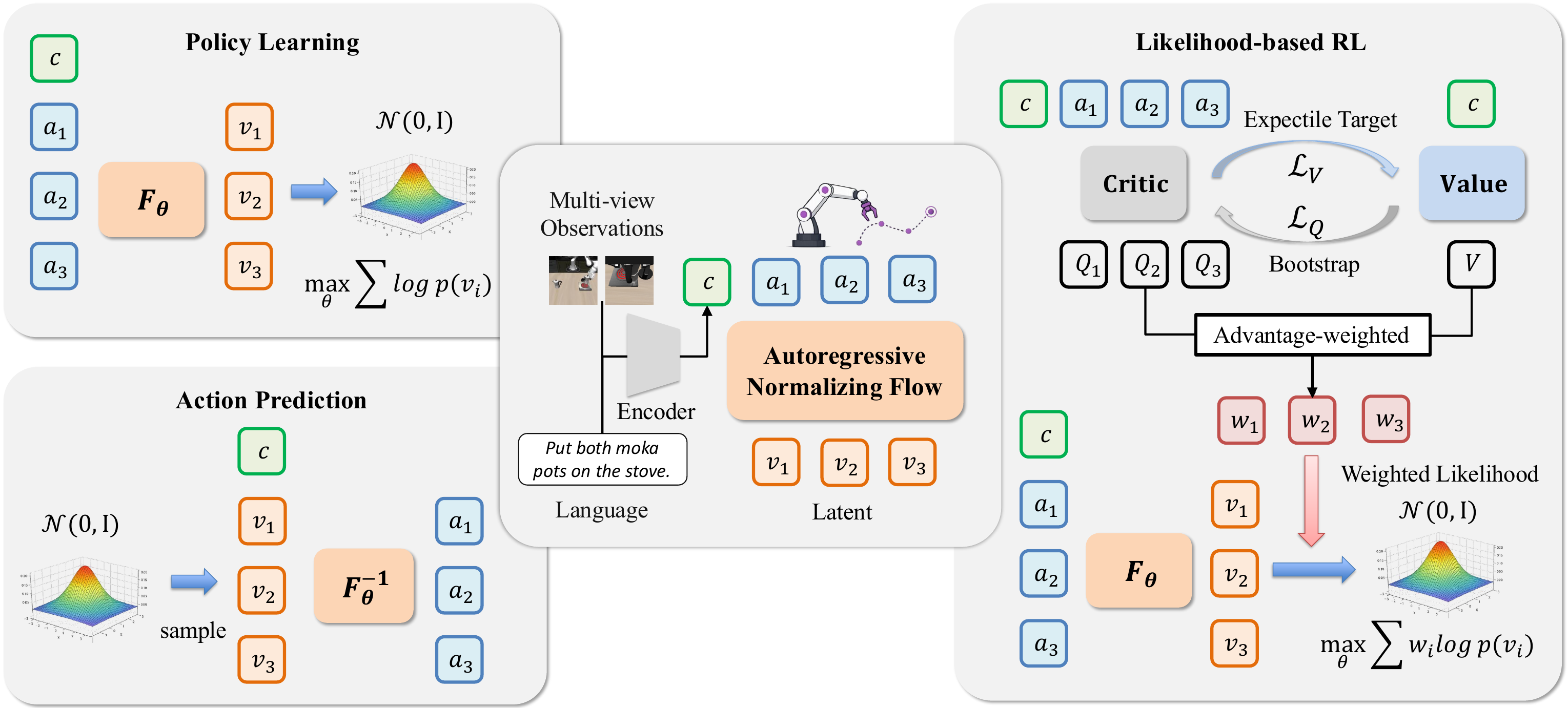}
\caption{
Overview of RoMAN-Flow.
RoMAN-Flow adopts an invertible AR-NF architecture that maps continuous action chunks to latent variables conditioned on multimodal inputs.
For imitation learning, it maximizes the exact conditional likelihood of demonstrated action chunks under a Gaussian latent prior.
For action generation, it samples latent variables from the Gaussian prior and recovers action chunks through the autoregressive inverse transformation.
For likelihood-based reinforcement learning, NF-IQL directly increases the exact likelihoods of high-advantage offline action chunks without autoregressive policy sampling.
}
\label{fig:method}
\end{figure*}

We develop a likelihood-based offline robotic reinforcement learning framework, named RoMAN-Flow, based on autoregressive normalizing flow (AR-NF) policies. This section starts with the problem setup and the construction of the conditional AR-NF policy. Then, we elaborate on NF-IQL post-training, which directly increases the exact likelihood of high-value actions from the offline dataset. After that, we detail the one-step policy distillation for efficient deployment. An overview of our framework is shown in Figure~\ref{fig:method}.

\subsection{Problem Setup}
\label{sec:problem_formulation}

We consider imitation learning followed by offline reinforcement learning post-training.
Given a fixed dataset of reward-labeled trajectories, we organize each trajectory into overlapping length-$H$ chunk transitions
\begin{equation}
x_{n,t}
=
\left(
c_{n,t},
\mathbf{a}_{n,t},
\mathbf{r}_{n,t},
c_{n,t+H}
\right),
\ 
\mathcal{D}_H
=
\left\{
x_{n,t}
\right\}_{n,t},
\label{eq:chunk_dataset}
\end{equation}
where $\mathbf{a}_{n,t}=(a_{n,t},\ldots,a_{n,t+H-1})$ is a continuous action chunk and $\mathbf{r}_{n,t}$ contains the rewards.
Imitation learning models the demonstrated action distribution, whereas offline RL uses the reward annotations to favor higher-value action chunks without additional environment interaction.
\subsection{Autoregressive Normalizing Flow Policy}
\label{sec:ar_nf_policy}
Given a chunk-level sample $(c_t,\mathbf{a}_t)$, RoMAN-Flow first encodes the visual observations, language instruction, and optional proprioceptive states into multimodal context tokens $\mathbf{C}_t$ using a pretrained encoder. Conditioned on context tokens $\mathbf{C}_t$, RoMAN-Flow then maps an observed action chunk $\mathbf{a}_t$ to a Gaussian latent $\mathbf{z}_t$ through an invertible AR-NF $F_\theta$.
The forward transformation provides exact conditional likelihoods for imitation learning and NF-IQL, whereas action generation applies the autoregressive inverse transformation $F_\theta^{-1}$ to a latent sampled from the Gaussian prior.

\paragraph{Multimodal context encoder.}
To fully exploit visual, linguistic, and proprioceptive conditions for action generation, we employ a pretrained vision-language model as the multimodal encoder.
Given the policy condition $c_t$, the encoder produces context tokens $\mathbf{C}_t$.
The context tokens are provided to every AR-NF flow block to condition action generation.
\paragraph{Autoregressive normalizing flow.}
We adopt SimFlow \cite{simflow} as the backbone of our conditional AR-NF model and adapt it to model continuous robotic action chunks.
The AR-NF transforms an action chunk $\mathbf{a}_t$ into a latent $\mathbf{z}_t$ through $L$ conditional invertible flow blocks $F_\theta
=
f_L
\circ
f_{L-1}
\circ
\cdots
\circ
f_1.$:
\begin{equation}
\mathbf{h}_t^{(0)}
=
\mathbf{a}_t,
\ 
\mathbf{h}_t^{(l)}
=
f_{l}
\left(
\mathbf{h}_t^{(l-1)};
\mathbf{C}_t
\right),
\ 
\mathbf{z}_t =\mathbf{h}_t^{(L)}
,
\label{eq:ar_nf_blocks}
\end{equation}
where $\mathbf{h}_t^{(l)}$ denotes the output of the $l$-th AR-NF block.

Each block employs a Transformer with a prefix-causal attention mask.
At action position $j$, the Transformer predicts an affine shift $\boldsymbol{\mu}_{t,j}^{(l)}$ and log-scale $\mathbf{s}_{t,j}^{(l)}$ using the multimodal context and preceding action positions:
\begin{equation}
\left(
\boldsymbol{\mu}_{t,j}^{(l)},
\mathbf{s}_{t,j}^{(l)}
\right)
=
T_l
\left(
\mathbf{C}_t,
\mathbf{h}_{t,<j}^{(l-1)}
\right).
\label{eq:ar_nf_parameters}
\end{equation}
The corresponding forward transformation is
\begin{equation}
\mathbf{h}_{t,j}^{(l)}
=
\left(
\mathbf{h}_{t,j}^{(l-1)}
-
\boldsymbol{\mu}_{t,j}^{(l)}
\right)
\odot
\exp
\left(
-\mathbf{s}_{t,j}^{(l)}
\right).
\label{eq:ar_nf_affine}
\end{equation}

Because the affine parameters at position $j$ depend only on $\mathbf{C}_t$ and preceding positions, each flow block has a triangular Jacobian.
The exact conditional log-likelihood is therefore
\begin{equation}
\log
\pi_\theta
\left(
\mathbf{a}_t
\mid
c_t
\right)
=
\log
p_0
\left(
\mathbf{z}_t
\right)
-
\sum_{l=1}^{L}
\sum_{j=1}^{H}
\mathbf{s}_{t,j}^{(l)},
\label{eq:ar_nf_likelihood}
\end{equation}
where $p_0(\mathbf{z})=\mathcal{N}(\mathbf{0},\mathbf{I})$ is the standard Gaussian density.

\paragraph{Imitation learning and action generation.}
We initialize RoMAN-Flow by minimizing the negative log-likelihood of demonstrated action chunks:
\begin{equation}
\mathcal{L}_{\mathrm{IL}}
=
-
\mathbb{E}_{(c_t,\mathbf{a}_t)\sim\mathcal{D}_H}
\left[
\log
\pi_\theta
\left(
\mathbf{a}_t
\mid
c_t
\right)
\right].
\label{eq:ar_nf_il}
\end{equation}
Because the complete action chunk is observed during training, the affine parameters for all positions can be evaluated in parallel through a single masked Transformer forward pass.
This enables efficient exact-likelihood optimization during both imitation learning and NF-IQL post-training.

At inference time, the policy samples
$\mathbf{z}_t\sim\mathcal{N}(\mathbf{0},\mathbf{I})$
and generates an action chunk through
\begin{equation}
\mathbf{a}_t
=
F_\theta^{-1}
\left(
\mathbf{z}_t;
\mathbf{C}_t
\right),
\qquad
F_\theta^{-1}
=
f_1^{-1}
\circ
f_2^{-1}
\circ
\cdots
\circ
f_L^{-1}.
\label{eq:ar_nf_inverse}
\end{equation}
Within each inverse block, the affine parameters depend on previously recovered action positions, requiring the action chunk to be reconstructed sequentially.
This forward--inverse asymmetry provides parallel exact-likelihood evaluation during training but introduces substantial latency at inference.

\subsection{NF-IQL Post-Training}
\label{sec:nf_iql}

To improve the behavior-cloned RoMAN-Flow policy beyond imitation learning, we develop \textbf{NF-IQL}, an offline reinforcement learning framework built upon IQL~\cite{kostrikov2021offline} and tailored to AR-NF policies.
Conventional actor-critic methods, such as TD3+BC~\cite{fujimoto2021minimalist}, require evaluating the critic on actions generated by the current policy during actor optimization.
For an AR-NF policy, generating these actions invokes the sequential inverse transformation and therefore incurs substantial training overhead.
NF-IQL instead performs policy improvement exclusively on action chunks contained in the offline dataset, avoiding current-policy action sampling.

NF-IQL consists of two coupled components.
First, it learns chunk-level action values and a state-value baseline from the offline data.
Second, it converts their difference into an advantage weight and uses this weight to optimize the exact conditional likelihood of each dataset action chunk.
The complete training procedure is summarized in the Appendix.

\paragraph{Chunk-level Q and state-value learning.}
We jointly learn an ensemble of $M$ Chunk Transformer critics
$\{Q_{\omega_m}\}_{m=1}^{M}$ and a state-value network $V_\phi$,
where $\omega_m$ and $\phi$ denote their trainable parameters. 
For stable value estimation, we maintain a target copy
$\{Q_{\bar{\omega}_m}\}_{m=1}^{M}$ of the online critic ensemble,
where $\bar{\omega}_m$ is updated from $\omega_m$ through Polyak averaging
and receives no gradient updates.
Given an offline action chunk, each critic $Q_{\omega_m}$ outputs
Q-value estimates for all action prefixes, while $V_\phi$ maps the
policy condition $c_t$ to a scalar state value $V_\phi(c_t)$.
The state value serves both as the bootstrap target for critic learning
and as the state-dependent baseline for subsequent policy improvement.

Following CO-RFT~\cite{huang2025co}, the $m$-th critic produces
prefix-level Q-value estimates in a single causally masked forward pass: $\{
Q_{\omega_m}^{(j)}
(
c_t,
a_{t:t+j}
)
\}_{j=0}^{H-1},
\ 
m=1,\ldots,M$,
where $j$ indexes the action prefix and
$a_{t:t+j}=(a_t,\ldots,a_{t+j})$.
Each action token attends to the policy condition and preceding actions while being prevented from accessing future actions.

For the $j$-th action prefix, we construct a temporal-difference target using the discounted rewards accumulated within the prefix and the bootstrapped value of the subsequent state:
\begin{equation}
y_{t,j}
=
\sum_{i=0}^{j}
\gamma^i r_{t+i}
+
\gamma^{j+1}
\operatorname{sg}
\left[
V_\phi
\left(
c_{t+j+1}
\right)
\right],
\label{eq:prefix_td_target}
\end{equation}
where $\gamma\in(0,1]$ denotes the discount factor and $\operatorname{sg}[\cdot]$ denotes the stop-gradient operation.
The online critics are trained by regressing all prefix-level Q-values toward their corresponding temporal-difference targets:
\begin{equation}
\mathcal{L}_{Q}
=
\mathbb{E}_{\mathcal{D}_{H}}
\left[
\frac{1}{MH}
\sum_{m=1}^{M}
\sum_{j=0}^{H-1}
\left(
Q_{\omega_m}^{(j)}
\left(
c_t,
a_{t:t+j}
\right)
-
y_{t,j}
\right)^2
\right].
\label{eq:prefix_critic_loss}
\end{equation}

To obtain a scalar value for the complete action chunk, we average the
target-critic estimates over both action prefixes and ensemble members:
\begin{equation}
\bar{Q}_{\bar{\omega}}
\left(
c_t,
\mathbf{a}_t
\right)
=
\frac{1}{MH}
\sum_{m=1}^{M}
\sum_{j=0}^{H-1}
Q_{\bar{\omega}_m}^{(j)}
\left(
c_t,
a_{t:t+j}
\right),
\label{eq:chunk_q_aggregation}
\end{equation}
where $\bar{\omega}_m$ denotes the parameters of the $m$-th target critic.

The state-value network is optimized through expectile regression over the aggregated target-critic estimates:
\begin{equation}
\mathcal{L}_{V}
=
\mathbb{E}_{(c_t,\mathbf{a}_t)\sim\mathcal{D}_{H}}
\left[
\rho_\tau
\left(
\operatorname{sg}
\left[
\bar{Q}_{\bar{\omega}}
\left(
c_t,
\mathbf{a}_t
\right)
\right]
-
V_\phi(c_t)
\right)
\right],
\label{eq:iql_value_loss}
\end{equation}
where $\rho_\tau(u)=|\tau-\mathbb{I}[u<0]|u^2$ and
$\tau\in(0,1)$ is the expectile parameter.
For $\tau>0.5$, this objective yields an upper-expectile state-value baseline.

\paragraph{Advantage-weighted likelihood optimization.}
Using the aggregated target-critic estimate and the state-value baseline, we define the advantage of each offline action chunk and its corresponding likelihood weight as
\begin{equation}
A_t
=
\bar{Q}_{\bar{\omega}}
\left(
c_t,
\mathbf{a}_t
\right)
-
V_\phi(c_t),
\qquad
w_t
=
\exp
\left(
\beta A_t
\right),
\label{eq:iql_advantage}
\end{equation}
where $\beta>0$ controls the strength of advantage weighting.
Action chunks whose estimated values exceed the state-value baseline receive larger weights, whereas lower-value action chunks contribute less to policy optimization.

We post-train the RoMAN-Flow actor by minimizing the advantage-weighted negative log-likelihood:
\begin{equation}
\mathcal{L}_{\pi}
=
-
\mathbb{E}_{(c_t,\mathbf{a}_t)\sim\mathcal{D}_{H}}
\left[
\operatorname{sg}
\left[
w_t
\right]
\log
\pi_\theta
\left(
\mathbf{a}_t
\mid
c_t
\right)
\right].
\label{eq:nf_iql_policy}
\end{equation}
The target-critic estimate, state value, advantage, and likelihood weight are treated as fixed during the actor update.
NF-IQL therefore shifts the policy toward higher-value behaviors without requiring current-policy action generation.

Training schedules, benchmark-specific reward relabeling, and target-network updates are detailed in the Appendix.

\begin{figure}
    \centering
    \includegraphics[width=0.8\linewidth]{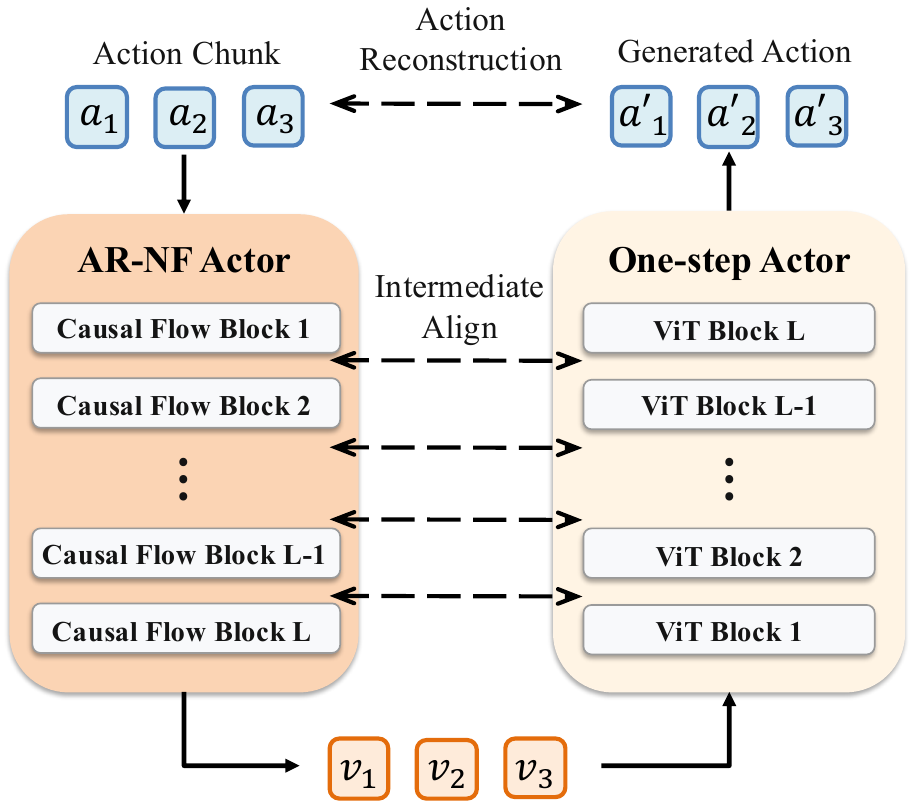}
    \caption{
One-step policy distillation.
The student model aligns with the teacher's intermediate reverse states and reconstructs the final action chunk in a single forward pass.
}
    \label{fig:distll}
\end{figure}
\subsection{One-Step Policy Distillation}
\label{sec:one_step_distillation}

Although NF-IQL avoids autoregressive action sampling during offline post-training, deploying the resulting AR-NF policy still requires sequential inverse transformation.
To accelerate policy execution, inspired by FARMER~\cite{farmer} and BiFlow~\cite{biflow}, we freeze the NF-IQL-post-trained RoMAN-Flow policy as a teacher and distill its autoregressive inverse mapping into a bidirectional student that generates an entire action chunk in one forward pass.

\paragraph{Data-induced trajectory distillation.}
As shown in Figure~\ref{fig:distll}, we instantiate the student $g_\psi$ as a bidirectional Transformer that processes all action positions in parallel.
Given an offline action chunk $\mathbf{a}_t$, we first apply a small Gaussian perturbation,
$\widetilde{\mathbf{a}}_t=\mathbf{a}_t+\boldsymbol{\epsilon}_t$ with
$\boldsymbol{\epsilon}_t\sim\mathcal{N}(\mathbf{0},\sigma^2\mathbf{I})$.
The frozen teacher maps $\widetilde{\mathbf{a}}_t$ to a latent code $\mathbf{z}_t$ and records the intermediate flow states
$\{\mathbf{h}_t^{(l)}\}_{l=0}^{L-1}$.
Conditioned on the same multimodal context $\mathbf{C}_t$, the student predicts the complete action chunk and its intermediate states:
\begin{equation}
\left(
\widehat{\mathbf{u}}_t^{(1)},
\ldots,
\widehat{\mathbf{u}}_t^{(L)},
\widehat{\mathbf{a}}_t
\right)
=
g_\psi
\left(
\mathbf{z}_t;
\mathbf{C}_t
\right).
\label{eq:student_reverse_trajectory}
\end{equation}

We train the student through intermediate-state alignment and action-chunk reconstruction:
\begin{equation}
\mathcal{L}_{\mathrm{data}}
=
\frac{\lambda_s}{L}
\sum_{r=1}^{L}
\left\|
\widehat{\mathbf{u}}_t^{(r)}
-
\mathbf{h}_t^{(L-r)}
\right\|_2^2
+
\lambda_a
\left\|
\widehat{\mathbf{a}}_t
-
\mathbf{a}_t
\right\|_2^2,
\label{eq:data_distillation}
\end{equation}
where $\lambda_s$ and $\lambda_a$ balance intermediate-state alignment and action reconstruction, respectively.

\paragraph{Prior-sampled trajectory distillation.}
The data-induced branch covers only latent codes obtained from offline action chunks.
To expose the student to a broader portion of the post-trained teacher distribution, we additionally introduce prior-sampled distillation.
For each policy condition, we sample latent $\mathbf{z}_t^{p}$
and decode it through the frozen NF-IQL teacher:
\begin{equation}
\mathbf{z}_t^{p}\sim\mathcal{N}(\mathbf{0},\mathbf{I}),
\ 
\left(
\mathbf{a}_t^{p},
\{\mathbf{h}_t^{p,(l)}\}_{l=0}^{L-1}
\right)
=
F_{\theta^\star}^{-1}
\left(
\mathbf{z}_t^{p};
\mathbf{C}_t
\right).
\label{eq:prior_teacher_trajectory}
\end{equation}
where $\theta^\star$ denotes the frozen teacher parameters.
Applying the same intermediate-state alignment and action-reconstruction objective to these teacher-generated trajectories defines the prior-sampling loss $\mathcal{L}_{\mathrm{prior}}$.

The complete distillation objective is
\begin{equation}
\mathcal{L}_{\mathrm{distill}}
=
\mathcal{L}_{\mathrm{data}}
+
\lambda_p
\mathcal{L}_{\mathrm{prior}},
\label{eq:distillation_objective}
\end{equation}
where $\lambda_p$ controls the contribution of the prior-sampling loss.
\section{Experiments}

\subsection{Experimental Setup}

\paragraph{Benchmarks.}
We evaluate RoMAN-Flow on MetaWorld-MT50~\cite{yu2020meta}, LIBERO~\cite{liu2023libero}, RoboMimic MH~\cite{mandlekar2021matters}, and a real-robot platform.
MetaWorld-MT50 contains 50 manipulation tasks and evaluates large-scale multitask learning; we report the average success rate across all four difficulty levels.
LIBERO consists of four ten-task suites: LIBERO-Spatial, LIBERO-Object, LIBERO-Goal, and LIBERO-Long, which evaluate generalization across spatial configurations, object identities, language-specified goals, and long-horizon behaviors, respectively.
On RoboMimic MH, we conduct a controlled comparison with SERNF~\cite{yang2026sernf}, which combines a conventional RealNVP policy with TD3+BC.

\begin{figure}[t]
\centering
\includegraphics[width=\columnwidth]{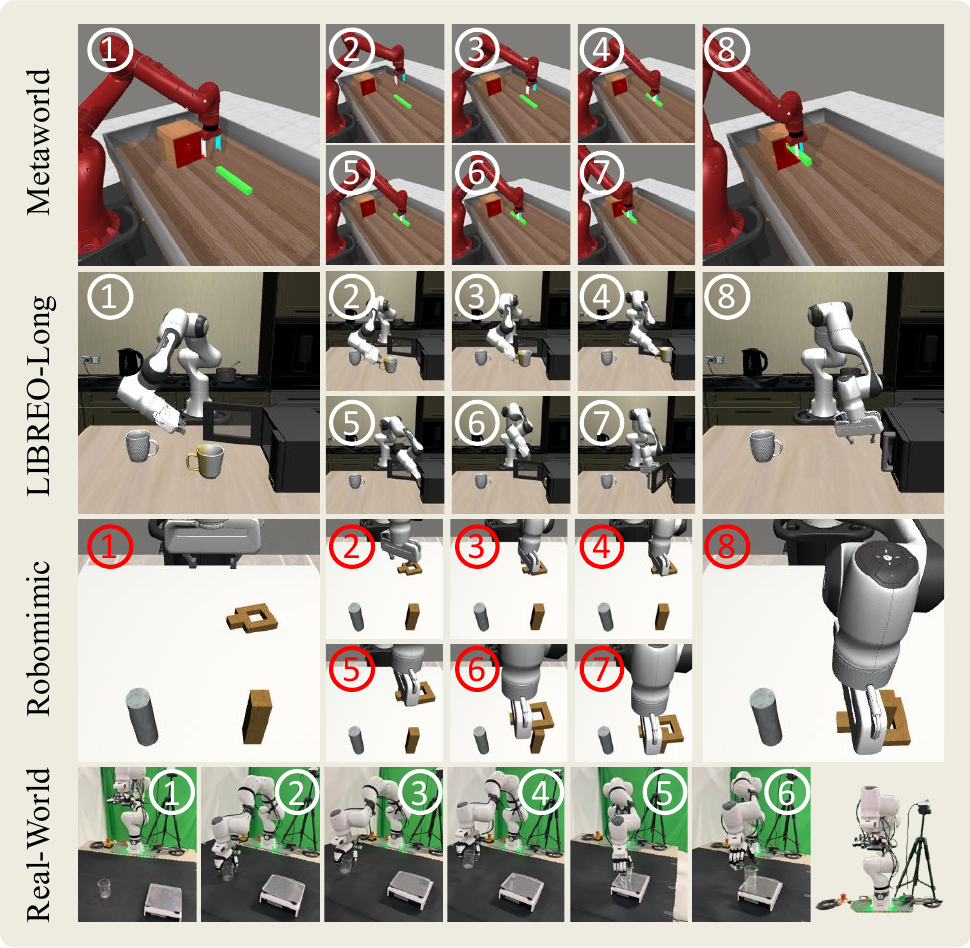}
\caption{
Qualitative results of RoMAN-Flow on MetaWorld, LIBERO-Long, RoboMimic, and real-world tasks.
Each row shows selected frames from a rollout, while the bottom-right inset presents our real-robot platform with a seven-DoF Franka arm and a twelve-DoF XHand dexterous hand.
}
\label{fig:franka-xhand-setup}
\end{figure}

\paragraph{Real-world evaluation.}
We evaluate RoMAN-Flow on the Franka--XHand platform shown in Figure~\ref{fig:franka-xhand-setup}, consisting of a seven-DoF Franka arm and a twelve-DoF XHand dexterous hand.
We consider four tasks: Pick Beaker, Pick Cylinder, Place Beaker, and Put Beaker on Balance.
At evaluation time, objects are initialized outside the spatial distribution covered by the offline demonstrations.
This setting assesses spatial out-of-distribution generalization and robustness to real-world visual variation, sensing noise, and control errors.

\begin{table*}[t]
\centering
{
\setlength{\tabcolsep}{4.2mm}
\begin{tabular}{lcccccc}
\toprule
Method & \#Params & Easy & Medium & Hard
& Very Hard & Average \\
\hline
Diffusion Policy~\cite{chi2025diffusion}       & 157M & 23.1 & 10.7 &  1.9 &  6.1 & 10.5 \\
TinyVLA~\cite{wen2025tinyvla}                &      1.3B & 77.6 & 21.5 & 11.4 & 15.8 & 31.6 \\
SmolVLA~\cite{shukor2025smolvla}                & 2.25B & \underline{87.1} & 51.8 & 70.0 & 64.0 & 68.2 \\
$\pi_0$~\cite{black2024pi0}                &     3.3B & 77.9 & 51.8 & 53.3 & 20.0 & 50.8 \\
$\pi_0$ + Flow-SDE~\cite{chen2025pirl}          & 3.3B & \textbf{92.1} & \textbf{74.6} & 61.7
                     & \textbf{84.0} & 78.1 \\
$\pi_{0.5}$~\cite{physicalintelligence2025pi05}    &  2.3B    & 68.2 & 37.3 & 41.7 & 28.0 & 43.8 \\
$\pi_{0.5}$ + Flow-SDE~\cite{chen2025pirl}      &  2.3B & 86.4 & 55.5 & 75.0 & 66.0 & 70.7 \\
\hline
\textbf{RoMAN-Flow (IL)}
                     & 1.45B & 80.4 & 63.6 & 75.0 & 72.0 & 72.8 \\
\textbf{RoMAN-Flow (NF-IQL)}
                     & 1.45B & 80.0 & \underline{74.5} & \textbf{90.0} & \underline{80.0} & \textbf{81.1} \\
\textbf{RoMAN-Flow (One-Step)}
                     & 0.56B & 81.4 & 72.7 & \underline{80.0} & \underline{80.0} & \underline{78.5} \\
\bottomrule
\end{tabular}
}
\caption{Success rates (\%) on MetaWorld-MT50. 
Avg. is the unweighted mean over the four difficulty groups. Bold and underlined numerical entries denote the best and second-best results, respectively.}
\label{tab:mt50_comparison}
\end{table*}

\begin{table*}[t]
\setlength{\tabcolsep}{5.2mm}
\centering
\begin{tabular}{lcccccc}
\toprule
Method & \#Params & Spatial & Object & Goal & Long & Average \\
\hline
Diffusion Policy~\cite{chi2025diffusion}
  & 157M
  & 78.3 & 92.5 & 68.3 & 50.5 & 72.4 \\
OpenVLA~\cite{kim2024openvla}
  & 7.5B
  & 84.7 & 88.4 & 79.2 & 53.7 & 76.5 \\
NORA-Long~\cite{hung2025nora}
  & 3B
  & 92.2 & 95.4 & 89.4 & 74.6 & 87.9 \\
SmolVLA~\cite{shukor2025smolvla}
  & 2.25B
  & 93.0 & 94.0 & 91.0 & 77.0 & 88.8 \\
GR00T-N1~\cite{bjorck2025gr00t}
  & 2.2B
  & 94.4 & 97.6 & 93.0 & 90.6 & 93.9 \\
$\pi_0$~\cite{black2024pi0}
  & 3.3B
  & \textbf{96.8} & 98.8 & \textbf{95.8}
  & 85.2 & 94.2 \\
UniVLA (Full)~\cite{bu2025univla}
  & $\sim$7.5B
  & \underline{96.5} & 96.8 & \underline{95.6}
  & 92.0 & \underline{95.2} \\
\hline
\textbf{RoMAN-Flow (IL)}
  & 1.45B
  & 95.0 & \underline{99.2} & 94.2
  & 85.6 & 93.5 \\
\textbf{RoMAN-Flow (NF-IQL)}
  & 1.45B
  & 95.0 & \textbf{99.4} & 94.6
  & \underline{92.2} & \textbf{95.3} \\
\textbf{RoMAN-Flow (One-Step)}
  & 0.56B
  & 94.4 & 95.6 & 91.8
  & \textbf{93.0} & 93.7 \\
\bottomrule
\end{tabular}
\caption{Success rates (\%) on the four LIBERO task suites. 
Average is the unweighted mean across suites. Bold and underlined numerical entries denote the best and second-best results, respectively.}
\label{tab:libero-benchmark-comparison}
\end{table*}

\begin{table*}[t]
\setlength{\tabcolsep}{3.5mm}
\centering
\begin{tabular}{lccccc}
\toprule
Method
& Pick\_beaker
& Pick\_cylinder
& Place\_beaker
& Balance
& Average \\
\hline
Diffusion Policy~\cite{chi2025diffusion}
& 6
& 0
& 0
& 0
& 1.5 \\
$\pi_{0}$~\cite{black2024pi0}
& 80
& 3
& 93
& 73
& 62.3 \\
$\pi_{0.5}$~\cite{physicalintelligence2025pi05}
& 93
& 10
& 97
& 87
& 71.8 \\
\hline
RoMAN-Flow (IL)
& {36}
& {20}
& \textbf{100}
& {73}
& {57.3} \\
RoMAN-Flow (NF-IQL)
& \textbf{100}
& \textbf{33}
& \textbf{100}
& {93}
& \textbf{81.5} \\
RoMAN-Flow (One-Step)
& {78}
& {20}
& {80}
& \textbf{100}
& {69.5} \\
\bottomrule
\end{tabular}
\caption{Success rates (\%) on four real-robot manipulation tasks. Average denotes the unweighted mean across the four tasks.}
\label{tab:real-robot-comparison}
\end{table*}

\begin{table}[t]
\centering
{\small
\setlength{\tabcolsep}{2.2mm}
\begin{tabular}{lccc}
\toprule
Method
& Lift
& Can
& Square \\
\hline
SERNF (IL)
& $79 \pm 0$
& $96 \pm 0$
& $61 \pm 4$ \\

SERNF (TD3+BC)
& $91 \pm 4$
& $96 \pm 2$
& $68 \pm 5$ \\
\hline
\textbf{RoMAN-Flow (IL)}
& $\mathbf{100 \pm 0}$
& $\mathbf{97 \pm 2}$
& $80 \pm 3$ \\

\textbf{RoMAN-Flow (NF-IQL)}
& $\mathbf{100 \pm 0}$
& $96 \pm 3$
& $\mathbf{85 \pm 3}$ \\

\textbf{RoMAN-Flow (One-Step)}
& $\mathbf{100 \pm 0}$
& $95 \pm 2$
& $80 \pm 3$ \\
\bottomrule
\end{tabular}
}
\caption{Success rates (\%) on RoboMimic MH. Results are reported as the mean and standard deviation over four training seeds, with 100 evaluation rollouts per seed.}
\label{tab:robomimic_mh_comparison}
\end{table}

\subsection{Main Results}

We compare three stages of RoMAN-Flow: imitation learning, denoted as RoMAN-Flow (IL); NF-IQL post-training, denoted as RoMAN-Flow (NF-IQL); and the distilled policy, denoted as RoMAN-Flow (One-Step).
All controlled comparisons use the same offline data and evaluation settings.
Training and evaluation settings are detailed in the Appendix.

\paragraph{Qualitative results.}
Figure~\ref{fig:franka-xhand-setup} presents representative successful rollouts from MetaWorld, LIBERO-Long, RoboMimic, and our real-world setup.
The sequences show coherent multi-stage behaviors, including approaching and manipulating objects, transporting them across the workspace, and completing precise target placement.
In particular, RoMAN-Flow successfully executes the real-world task of placing a beaker onto a balance.
Additional qualitative examples are provided in the Appendix.

\paragraph{Large-scale multitask evaluation on MetaWorld.}
Table~\ref{tab:mt50_comparison} reports success rates on MetaWorld-MT50 using the four difficulty groups and evaluation protocol defined by $\pi_{\mathrm{RL}}$~\cite{chen2025pirl}.
Compared with RoMAN-Flow (IL), NF-IQL increases the unweighted mean across difficulty groups from 72.8\% to 81.1\%, with gains concentrated in the Medium, Hard, and Very Hard groups.
RoMAN-Flow (One-Step) achieves 78.5\%, retaining most of the NF-IQL teacher's multitask performance while replacing sequential AR-NF inversion with one-step inference.
Under the aligned protocol, RoMAN-Flow (NF-IQL) also outperforms $\pi_0$ + Flow-SDE by 3.0 percentage points.
This comparison is particularly notable because $\pi_0$ + Flow-SDE uses online environment interaction, whereas NF-IQL operates exclusively on fixed offline data.

\paragraph{Multi-dimensional task generalization on LIBERO.}
Table~\ref{tab:libero-benchmark-comparison} reports results on the four official LIBERO suites.
NF-IQL increases the mean success rate from 93.5\% to 95.3\%, primarily through a 6.6-percentage-point improvement on LIBERO-Long, while preserving the strong IL performance on Spatial, Object, and Goal.
RoMAN-Flow (NF-IQL) achieves the highest average success rate among the compared methods while using fewer inference-time parameters than the listed large-scale VLA baselines.
The one-step student attains an average success rate of 93.7\% and achieves the highest result on LIBERO-Long, indicating that the distilled policy remains effective on long-horizon manipulation.

\paragraph{Comparison with SERNF on RoboMimic.}
Table~\ref{tab:robomimic_mh_comparison} compares RoMAN-Flow with SERNF~\cite{yang2026sernf} under the training and evaluation protocol used by SERNF.
At the imitation-learning stage, RoMAN-Flow matches or exceeds SERNF on all three tasks, supporting the effectiveness of AR-NFs for continuous-action policy modeling.
After offline RL post-training, RoMAN-Flow (NF-IQL) achieves success rates of 100\%, 96\%, and 85\% on Lift, Can, and Square, respectively, compared with 91\%, 96\%, and 68\% for SERNF (TD3+BC).
The one-step student retains strong performance, achieving 100\%, 95\%, and 80\% on the three tasks.

\paragraph{Real-world manipulation.}
Table~\ref{tab:real-robot-comparison} reports success rates on four real-robot manipulation tasks.
NF-IQL increases the average success rate from 57.3\% to 81.5\%, corresponding to an absolute improvement of 24.2 percentage points.
The largest gains occur on Pick Beaker, Pick Cylinder, and Put Beaker on Balance, where success rates improve by 64, 13, and 20 percentage points, respectively.
These results show that offline RL post-training can effectively use reward information to refine grasping and precise placement behaviors under real-world visual and control variability.
RoMAN-Flow (NF-IQL) further outperforms these baselines across all four tasks, achieving an average success rate of 81.5\%, a 9.7-percentage-point improvement over the stronger $\pi_{0.5}$ baseline.
\begin{figure}[t]
\centering
\includegraphics[width=\columnwidth]{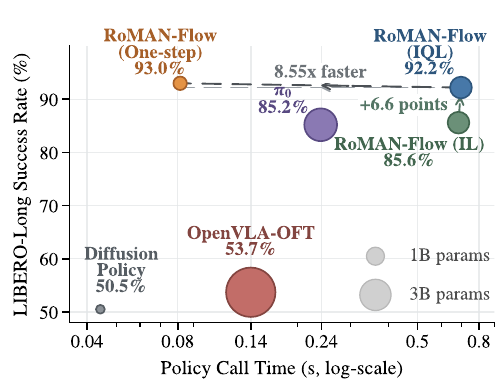}
\caption{Speed--performance trade-off on LIBERO-Long with action horizon $H=16$. Each point reports the average success rate and action-chunk generation latency over the ten LIBERO-Long tasks, with marker area proportional to the number of inference-time policy parameters.}
\label{fig:libero10-speed-performance}
\end{figure}
\subsection{One-Step Distillation and Efficient Inference}
\label{sec:one_step_results}

RoMAN-Flow (NF-IQL) incurs high inference latency because its autoregressive inverse transformation generates action chunks sequentially.
We therefore distill it into a bidirectional one-step student.
Across the simulation benchmarks, the student achieves performance comparable to its NF-IQL teacher, slightly outperforming it on LIBERO-Long and retaining approximately 97\% of its performance on MetaWorld-MT50 and RoboMimic MH.
As shown in Figure~\ref{fig:libero10-speed-performance}, the student achieves a LIBERO-Long success rate of 93.0\%, compared with 92.2\% for the teacher, while reducing action-chunk generation latency from approximately 697\,ms to 81.5\,ms, corresponding to an $8.55\times$ speedup.

\subsection{Ablation Study}
\label{sec:ablations}

\paragraph{Effect of AR-NF actor capacity.}
We study the scaling behavior of the AR-NF actor on RoboMimic Square-MH. Holding the visual encoder, offline dataset, and training and evaluation protocol fixed, we vary only the width, number of attention heads, and depth of the AR-NF actor. As shown in Table~\ref{tab:robomimic-nf-scale-ablation}, the S, B, and L configurations span 33.8M to 466.2M parameters but achieve nearly identical success rates of 77\%--78\%. This suggests that increasing capacity within the small-to-medium scale regime does not yet translate into improved modeling of fine-grained action structure. Once scaled to 685.5M parameters, the XL configuration reaches an 85\% success rate. These results indicate that sufficiently large actor capacity can better capture the complex, precision-sensitive action distributions required by Square-MH.

\begin{table}[t]
\centering
{
\setlength{\tabcolsep}{1.7mm}
\begin{tabular}{lccccc}
\toprule
Size & Dim. & Heads & Layer Depth & \#Params & SR \\
\hline
S  & 384  & 6  & $[2,2,2,2,2,2]$    & 33.8M  & 78.0 \\
B  & 768  & 12 & $[2,2,2,2,2,2]$    & 134.8M & 77.0 \\
L  & 1024 & 16 & $[2,2,2,2,2,14]$ & 466.2M & 78.0 \\
XL & 1152 & 16 & $[2,2,2,2,2,18]$ & 685.5M & \textbf{85.0} \\
\bottomrule
\end{tabular}
}
\caption{The capacity ablation of AR-NF actor on RoboMimic Square-MH. 
SR denotes success rate (\%).}
\label{tab:robomimic-nf-scale-ablation}
\end{table}

\section{Conclusion}

We introduced RoMAN-Flow, a likelihood-based offline reinforcement learning framework tailored for autoregressive normalizing flow policies in robot manipulation.
RoMAN-Flow addresses the two main challenges of AR-NFs: NF-IQL avoids costly autoregressive sampling during post-training, while one-step distillation removes sequential inversion at deployment.
Experiments on simulation and real-robot benchmarks demonstrate effective policy improvement and an $8.55\times$ inference speedup with limited performance loss.
Our work establishes AR-NFs as a practical policy family for likelihood-based offline robotic reinforcement learning.

\section*{Acknowledgments}
We would like to extend our deepest appreciation to Qinyu Zhao  for insightful discussions.

\bibliography{aaai2027}
\clearpage

\section{Implementation Details}
\subsection{Training Pipeline}

The training pipeline is summarized in Algorithm~\ref{alg:arnf_training}.
We first train the AR-NF policy on offline action chunks through maximum-likelihood imitation learning.
NF-IQL post-training then proceeds in two phases: we first freeze the actor and warm up the prefix-level critics and state-value function, and subsequently update the critics, value function, and actor using advantage-weighted likelihood optimization over offline action chunks.
Finally, we freeze the post-trained AR-NF policy as the teacher and train a one-step student to approximate its inverse mapping through intermediate-state alignment and action-chunk reconstruction, optionally augmented with teacher trajectories generated from prior-sampled latents.
The pipeline outputs the post-trained AR-NF policy and the distilled one-step student.

\begin{algorithm}[!t]
\caption{RoMAN-Flow Training and Distillation}
\label{alg:arnf_training}

\textbf{Input}: Offline dataset $\mathcal{D}$ \\
\textbf{Output}: Post-trained policy $\pi_{\theta^\star}$ and
one-step student $g_{\psi^\star}$

\begin{algorithmic}[1]
\setlength{\itemsep}{0pt}

\STATE Initialize $\pi_\theta$,
$\{Q_{\omega_k},Q_{\bar{\omega}_k}\}_{k=1}^{M}$, and $V_\phi$.

\STATE \textbf{Stage I: Imitation learning}
\FOR{$n=1$ to $N_{\mathrm{IL}}$}
    \STATE Sample $(c_t,\mathbf{a}_t)\sim\mathcal{D}$.
    \STATE Update $\theta$ with
    $\mathcal{L}_{\mathrm{IL}}
    =-\mathbb{E}[\log\pi_\theta(\mathbf{a}_t\mid c_t)]$.
\ENDFOR

\STATE \textbf{Stage II: NF-IQL post-training}
\STATE Freeze $\pi_\theta$ during value warm-up.
\FOR{$n=1,\ldots,N_{\mathrm{warm}}$}
    \STATE Sample $\mathcal{B}\sim\mathcal{D}$ and construct
    prefix targets $y_{t,j}$.
    \STATE Update the critics and value function using
    $\mathcal{L}_{Q}$ and $\mathcal{L}_{V}$.
    \STATE Polyak-update the target critics.
\ENDFOR

\STATE Unfreeze $\pi_\theta$.
\FOR{$n=1$ to $N_{\mathrm{IQL}}$}
    \STATE Update the critics and value function.
    \STATE Compute
    $A_t=\bar{Q}_{\bar{\omega}}(c_t,\mathbf{a}_t)-V_\phi(c_t)$.
    \STATE Update $\theta$ using
    $\mathcal{L}_{\pi}
    =-\mathbb{E}[
    e^{\beta A_t}
    \log\pi_\theta(\mathbf{a}_t\mid c_t)]$.
\ENDFOR

\STATE \textbf{Stage III: One-step distillation}
\STATE Freeze $\pi_{\theta^\star}$ and initialize $g_\psi$.
\FOR{$n=1$ to $N_{\mathrm{align}}$}
  \STATE Sample $(c_t,\mathbf{a}_t)\sim\mathcal{D}$.
  \STATE Run the teacher forward on $\mathbf{a}_t$ to obtain $\mathbf{z}_t$ and $\{\mathbf{h}_t^{(l)}\}_{l=0}^{L-1}$.
    \STATE 
  Run $g_\psi(\mathbf{z}_t;c_t)$ to obtain $\{\widehat{\mathbf{h}}_{t}^{(l)}\}_{l=0}^{L-1}$ and
  $\widehat{\mathbf{a}}_t$
  \STATE Sample $\mathbf{z}_t^{p}\sim p_0$ \STATE 
  Run the teacher inverse under $c_t$ to obtain $\{\mathbf{u}_{t,p}^{(l)}\}_{l=0}^{L-1}$ and $\mathbf{a}_t^{p}$.
  \STATE 
  Run $g_\psi(\mathbf{z}_t^{p};c_t)$ to obtain $\{\widehat{\mathbf{u}}_{t,p}^{(l)}\}_{l=0}^{L-1}$ and
  $\widehat{\mathbf{a}}_t^{p}$.
  \STATE
  Compute $\mathcal{L}_{\mathrm{data}}$ using $\{\widehat{\mathbf{h}}_{t}^{(l)}\}_{l=0}^{L-1}$ and $\{\mathbf{h}_t^{(l)}\}_{l=0}^{L-1}$, $\widehat{\mathbf{a}}_t$ and $\mathbf{a}_t$.
  
  \STATE Compute $\mathcal{L}_{\mathrm{prior}}$ using $\{\widehat{\mathbf{u}}_{t,p}^{(l)}\}_{l=0}^{L-1}$ and $\{\mathbf{u}_{t,p}^{(l)}\}_{l=0}^{L-1}$, $\widehat{\mathbf{a}}_t^{p}$ and $\mathbf{a}_t^{p}$.
  \STATE Update $\psi$ using
$\mathcal{L}_{\mathrm{data}}
+\lambda_p\mathcal{L}_{\mathrm{prior}}$.
\ENDFOR

\STATE \textbf{return}
$\pi_{\theta^\star}$ and $g_{\psi^\star}$.

\end{algorithmic}
\end{algorithm}
\subsection{Reward Relabeling}
To alleviate the credit-assignment difficulty caused by sparse rewards, we adopt trajectory-level reward relabeling inspired by HUBL~\cite{geng2023improving}. For trajectory $n$, we define the terminal-aware discount as $\gamma_{n,t}=\gamma(1-d_{n,t})$, where $d_{n,t}$ denotes the terminal indicator, and compute the Monte Carlo return as
\begin{equation}
G_{n,t}
=
r_{n,t}
+
\gamma_{n,t}G_{n,t+1}.
\end{equation}
Let $T_n$ denote the number of transitions in trajectory $n$. We define its trajectory score as the mean Monte Carlo return,
\begin{equation}
s_n
=
\frac{1}{T_n}
\sum_{t=0}^{T_n-1} G_{n,t},
\end{equation}
and determine a trajectory-level mixing coefficient according to its empirical rank in the offline dataset:
\begin{equation}
\lambda_n
=
\alpha\frac{1}{N}
\sum_{m=1}^{N}
\mathbb{I}[s_m\leq s_n],
\end{equation}
where $\alpha$ controls the relabeling strength. We then define the relabeled reward and bootstrap discount as
\begin{equation}
\widetilde{r}_{n,t}
=
r_{n,t}
+
\gamma_{n,t}\lambda_nG_{n,t+1},
\qquad
\widetilde{\gamma}_{n,t}
=
\gamma_{n,t}(1-\lambda_n).
\label{eq:hubl_relabel}
\end{equation}
This formulation adaptively interpolates between the observed Monte Carlo return and value-function bootstrapping, assigning greater reliance on observed future returns to higher-return trajectories. For action chunks, we recursively accumulate the relabeled rewards and discounts to construct the TD target for each action prefix.

\section{Experimental Details}
\label{app:experimental_details}

\subsection{Model and Training Configurations}
\label{app:model_training_details}

RoMAN-Flow is trained in three stages: imitation learning, NF-IQL post-training, and one-step distillation.
The complete optimization procedure is summarized in Algorithm~\ref{alg:arnf_training}; here, we describe the benchmark-specific model and training configurations.

\paragraph{Backbones and observations.}
For LIBERO, MetaWorld-MT50, and the real-robot experiments, we use SmolVLM-500M-Instruct as the multimodal encoder for visual observations, language instructions, and proprioceptive states.
The suite-specific LIBERO policies are initialized from a shared checkpoint pretrained on LIBERO-90 for 100k steps.
For RoboMimic MH, we follow the perception setup of SERNF~\cite{yang2026sernf} for a controlled comparison, replacing the VLM with two separate ImageNet-pretrained ResNet-18 encoders for the third-person and wrist-camera RGB observations.

\paragraph{Model sizes.}
The SmolVLM-based AR-NF policy used for imitation learning and NF-IQL contains approximately 1.45B inference-time parameters, while the corresponding one-step student contains approximately 0.56B parameters.
The ResNet-based AR-NF policy and one-step student used on RoboMimic contain approximately 0.95B and 78.2M parameters, respectively.
Table~\ref{tab:implementation-configurations} summarizes the model architectures and training hyperparameters.

\begin{table*}[t]
\centering
{\small
\setlength{\tabcolsep}{1.5mm}
\begin{tabular}{@{}lccccc@{}}
\toprule
& \multicolumn{4}{c}{\textbf{RoMAN-Flow (IL / NF-IQL)}}
& \textbf{RoMAN-Flow (One-Step)} \\
\cmidrule(lr){2-5}
\cmidrule(lr){6-6}
\textbf{Setting}
& \textbf{LIBERO}
& \textbf{MT50}
& \textbf{RoboMimic}
& \textbf{Real Robot}
& \textbf{Default} \\
\midrule

\multicolumn{6}{@{}l}{\textbf{Model configuration}} \\
\midrule

Hidden dimension
& 1152
& 1152
& 1152
& 1152
& 512 \\

Attention heads
& 16
& 16
& 16
& 16
& 8 \\

Flow / ViT blocks
& 6
& 6
& 6
& 6
& 7 \\

Inference parameters
& 1.45B
& 1.45B
& 0.95B
& 1.45B
& 0.56B / 78.2M \\

\midrule
\multicolumn{6}{@{}l}{\textbf{Training configuration}} \\
\midrule

Training steps
& 20k / 50k
& 100k / 60k
& 30k / 50k
& 50k / 30k
& 20k \\

Global batch size
& 64 / 64
& 128 / 128
& 144 / 144
& 32 / 32
& 64 \\

Actor LR
& $5{\times}10^{-5} / 2{\times}10^{-5}$
& $1{\times}10^{-4} / 5{\times}10^{-6}$
& $1{\times}10^{-4} / 2{\times}10^{-4}$
& $5{\times}10^{-5} / 1{\times}10^{-5}$
& -- \\

Student LR
& --
& --
& --
& --
& $1{\times}10^{-4}$ \\

Critic LR
& -- / $1{\times}10^{-4}$
& -- / $5{\times}10^{-5}$
& -- / $2{\times}10^{-4}$
& -- / $5{\times}10^{-5}$
& -- \\

Expectile $\tau$
& -- / 0.80
& -- / 0.75
& -- / 0.75
& -- / 0.75
& -- \\

Advantage temperature $\beta$
& -- / 10
& -- / 45
& -- / 10
& -- / 3
& -- \\

Advantage clip
& -- / 100
& -- / 5
& -- / 20
& -- / 20
& -- \\

HUBL coefficient
& -- / 0.20
& -- / 0
& -- / 0.15
& -- / 0
& -- \\

Critic warm-up
& -- / 1k
& -- / 5k
& -- / 5k
& -- / 1k
& -- \\

Discount $\gamma$
& -- / 0.995
& -- / 0.995
& -- / 0.997
& -- / 0.997
& -- \\

Target update rate
& -- / 0.005
& -- / 0.005
& -- / 0.05
& -- / 0.005
& -- \\

Prior-loss weight
& --
& --
& --
& --
& 1 \\

Prior-sample fraction
& --
& --
& --
& --
& 1 \\

Prior-sampling temperature
& --
& --
& --
& --
& 0.5 \\

EMA decay
& --
& --
& --
& --
& 0.9999 \\

Optimizer
& Adam
& Adam
& Adam
& Adam
& Adam \\

Weight decay
& 0
& 0
& 0
& 0
& 0 \\

Gradient clipping
& 1.0
& 1.0
& 1.0
& 1.0
& 1.0 \\

\bottomrule
\end{tabular}
}
\caption{
Model configurations and training hyperparameters.
Paired entries under RoMAN-Flow (IL / NF-IQL) report the IL and NF-IQL settings, respectively, while single entries are shared by both stages.
The One-Step column reports the default distillation configuration, with its two parameter counts corresponding to the SmolVLM- and ResNet-based students.
A dash denotes an inapplicable setting.
}
\label{tab:implementation-configurations}
\end{table*}

\subsection{Evaluation Protocols}
\label{app:evaluation_protocols}

All policies are evaluated from fixed checkpoints without parameter updates.
We use episode-level success rate as the primary metric and execute each predicted action chunk completely before replanning.

\paragraph{LIBERO.}
We follow the official LIBERO protocol~\cite{liu2023libero} in terms of task suites, success criteria, and initial states.
We evaluate LIBERO-Spatial, LIBERO-Object, LIBERO-Goal, and LIBERO-Long, each containing ten tasks.
Each task is evaluated over 50 episodes, resulting in 500 episodes per suite.
All policies predict and execute 16-step action chunks.

\paragraph{MetaWorld-MT50.}
Following the evaluation protocol of $\pi_{\mathrm{RL}}$~\cite{chen2025pirl}, we evaluate each of the 50 tasks over ten episodes using four parallel environments and a maximum episode length of 200 steps.
All policies predict and execute 5-step action chunks.
We report the unweighted mean success rate across the Easy, Medium, Hard, and Very Hard groups.

\paragraph{RoboMimic MH.}
We follow the training and evaluation protocol of SERNF~\cite{yang2026sernf} on Lift, Can, and Square.
For each of four training seeds, we evaluate the policy on the same 100 fixed initial states using episode seeds 0--99, yielding 400 episodes per task and training stage.
Each evaluation uses a single environment, a maximum episode length of 500 steps, and 10-step action chunks.

\paragraph{Real-robot evaluation.}
For real-robot evaluation, we use a dedicated GPU inference server and a separate robot-control workstation. Target objects are initialized within predefined evaluation regions that include locations outside the initial-position distribution represented in the offline demonstrations. The resulting protocol therefore covers both in-distribution and spatial out-of-distribution initializations. Each policy predicts 16-step action chunks.

\section{Benchmarks and Qualitative Examples}
\subsection{Metaworld}
\begin{figure*}[t]
\centering
\includegraphics[width=\textwidth]{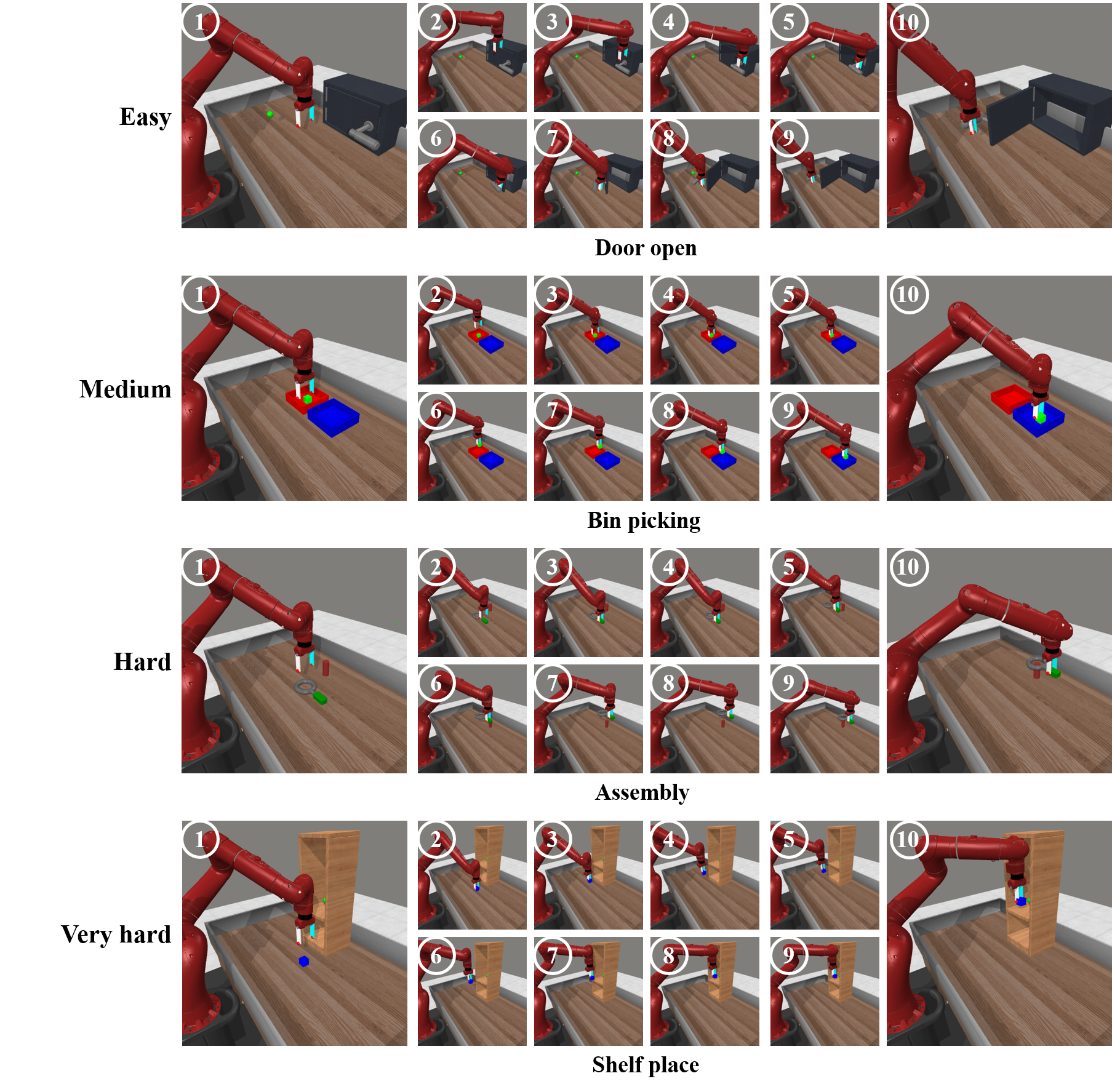}
\caption{Qualitative results of RoMAN-Flow on representative MetaWorld-MT50 manipulation tasks.}
\label{fig:metaworld-qualitative}
\end{figure*}

MetaWorld-MT50 is a public multitask offline manipulation dataset released by LeRobot and built upon the MetaWorld simulation benchmark~\cite{yu2020meta}. It covers 50 tabletop manipulation tasks performed by a Sawyer robot, spanning button pressing, articulated- object interaction, object grasping and transport, precise assembly, and long-horizon placement. The dataset contains 2,500 expert trajectories and 204,806 transitions, corresponding to 50 trajectories per task. Each sample provides a single-view RGB observation, a 4- dimensional robot state, a 4-dimensional continuous Cartesian action, and a natural-language task description. Following the task partition adopted by $\pi_{\mathrm{RL}}$~\cite{chen2025pirl}, the 50 tasks are divided into Easy, Medium, Hard, and Very Hard groups containing 28, 11, 6, and 5 tasks, respectively. Figure~\ref{fig:metaworld-qualitative} visualizes representative rollouts from the four difficulty groups: Door Open, Bin Picking, Assembly, and Shelf Place, illustrating articulated-object interaction, cross-bin pick-and-place, precise assembly, and long-horizon elevated placement.

\subsection{LIBERO}
\begin{figure*}[t]
\centering
\includegraphics[width=0.95\textwidth]{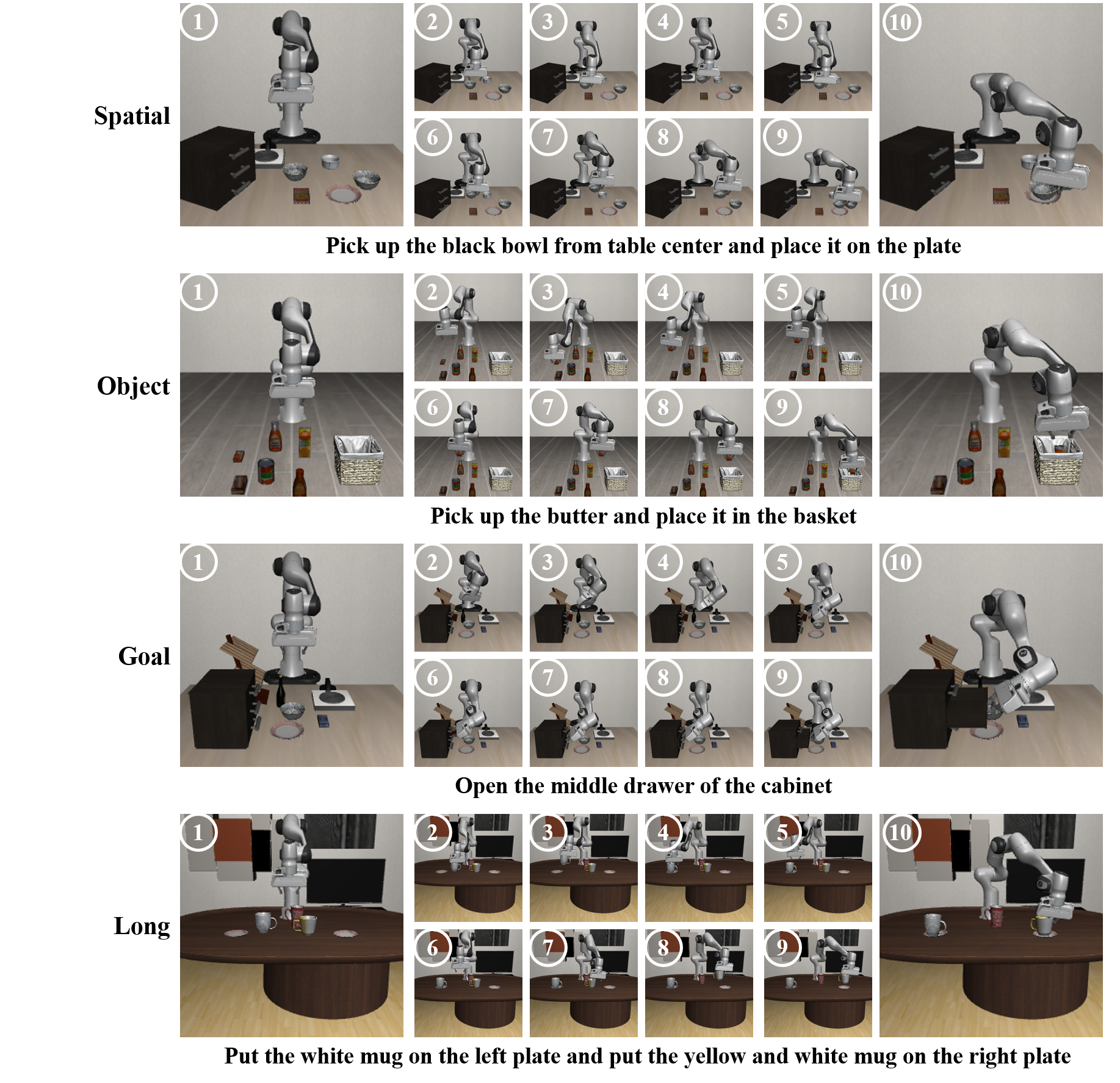}
\caption{Qualitative rollouts of RoMAN-Flow on representative tasks from the LIBERO benchmark.}
\label{fig:libero-qualitative}
\end{figure*}

LIBERO~\cite{liu2023libero} is a public language-conditioned manipulation benchmark designed to study knowledge transfer in multitask and lifelong robot learning. The complete benchmark contains 130 tabletop manipulation tasks performed by a Franka Panda robot. Following the common evaluation convention in recent VLA studies, we consider four ten-task suites: LIBERO-Spatial, LIBERO-Object, LIBERO-Goal, and LIBERO-Long, where LIBERO-Long corresponds to LIBERO-10 in the original benchmark. Spatial, Object, and Goal isolate knowledge transfer across spatial layouts, object identities, and language-specified goals, respectively, whereas Long contains longer-horizon tasks involving multiple consecutive manipulation stages. Each task provides 50 human-teleoperated demonstrations, resulting in 2,000 trajectories across the four evaluation suites. RoMAN-Flow uses two RGB views, an 8-dimensional proprioceptive state, and a language instruction. The suite-specific policies are initialized from a shared checkpoint pretrained on the 90 tasks and 4,500 demonstrations of LIBERO-90. Figure~\ref{fig:libero-qualitative} presents representative RoMAN-Flow rollouts across the LIBERO suites.

\subsection{Robomimic}
\begin{figure*}[t]
\centering
\includegraphics[width=0.95\textwidth]{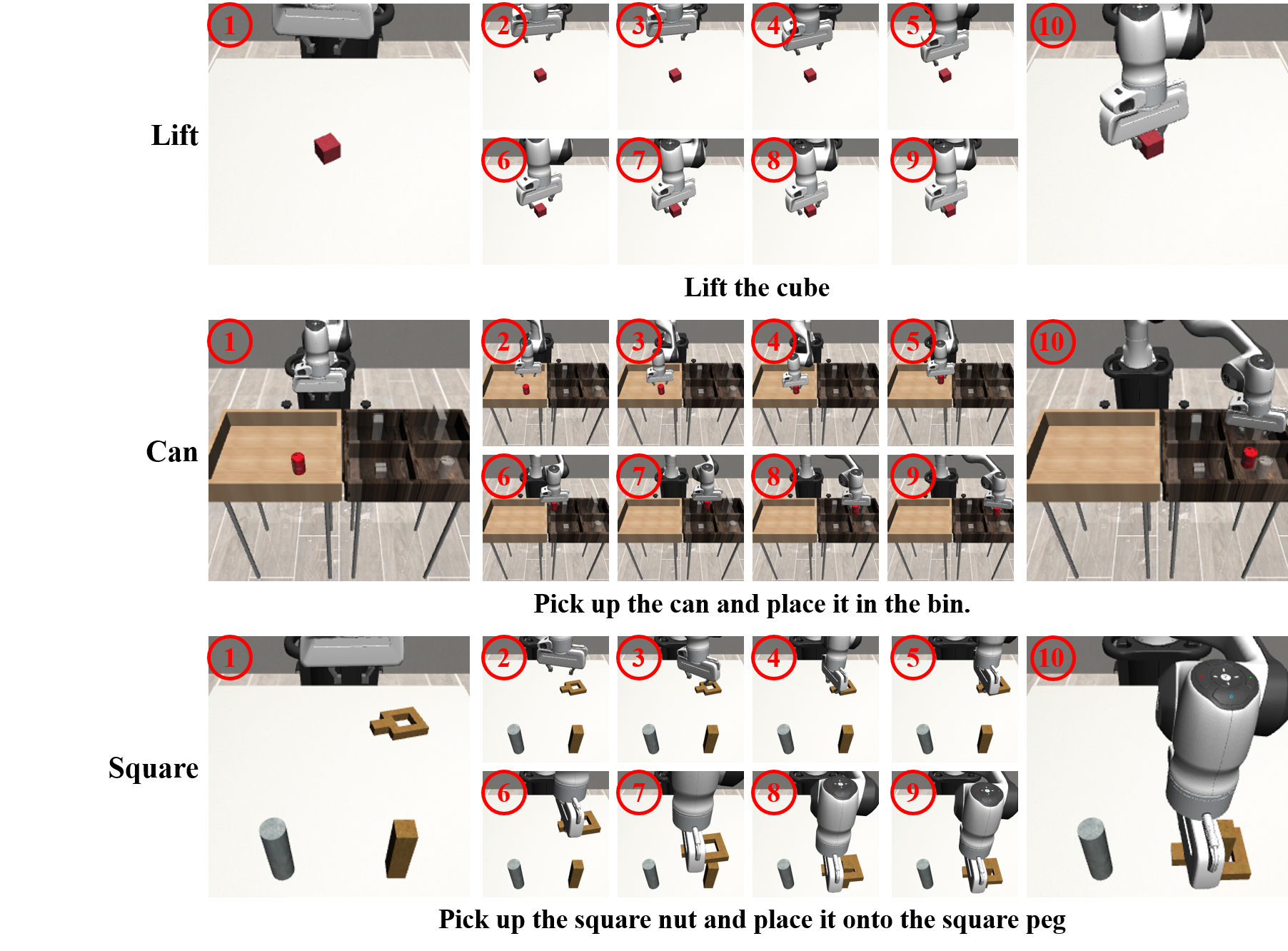}
\caption{Qualitative rollouts of RoMAN-Flow on the Lift, Can, and Square tasks from RoboMimic MH.}
\label{fig:robomimic-qualitative}
\end{figure*}

RoboMimic~\cite{mandlekar2021matters} is a public benchmark for learning robot manipulation policies from offline demonstrations. Its Multi-Human (MH) datasets contain behaviorally diverse demonstrations collected from six human operators with different levels of proficiency. We consider three tasks: Lift, which requires grasping and lifting a cube; Can, which requires transferring a can into a target bin; and Square, which requires precisely placing a square nut onto its corresponding peg. Each task provides 300 demonstrations, resulting in 900 trajectories and 174,614 transitions across the three tasks. To enable a controlled comparison with SERNF~\cite{yang2026sernf}, RoMAN-Flow does not use a VLM or proprioceptive inputs in these experiments. Instead, it observes two $84\times84$ RGB views from a third-person camera and a wrist-mounted camera and predicts 7-dimensional continuous control commands. The three tasks impose progressively greater manipulation demands, with Square requiring a longer sequence of precise grasping, alignment, and insertion behaviors. Figure~\ref{fig:robomimic-qualitative} presents representative RoMAN-Flow rollouts on Lift, Can, and Square.

\end{document}